\documentclass{article}

\usepackage{arxiv}
\usepackage{natbib}

\usepackage[utf8]{inputenc} 
\usepackage[T1]{fontenc}    
\usepackage{hyperref}       
\usepackage{url}            
\usepackage{booktabs}       
\usepackage{amsmath}        
\usepackage{amsfonts}       
\usepackage{nicefrac}       
\usepackage{microtype}      
\usepackage{lipsum}
\usepackage{graphicx}
\graphicspath{ {./images/} }

\title{From Stochastic Answers to Verifiable Reasoning: Interpretable Decision-Making with LLM-Generated Code}

\author{\mdseries
  Anirudh Jaidev Mahesh$^{1}$, 
  Ben Griffin$^{2}$, 
  Fuat Alican$^{3}$, 
  Joseph Ternasky$^{3}$, 
  Zakari Salifu$^{3}$, 
  Kelvin Amoaba$^{3}$, \\
  Yagiz Ihlamur$^{4}$,
  Aaron Ontoyin Yin$^{3}$, 
  Aikins Laryea$^{3}$,
  Afriyie Samuel$^{3}$, 
  Yigit Ihlamur$^{3}$
  \\[1em]
  $^{1}$Georgia Institute of Technology \quad
  $^{2}$University of Oxford \quad
  $^{3}$Vela Research
  $^{4}$Amazon
}
\begin{document}

\maketitle
\begin{abstract}
Large language models (LLMs) are increasingly used for high-stakes decision-making, yet existing approaches struggle to reconcile scalability, interpretability, and reproducibility. Black-box models obscure their reasoning, while recent LLM-based rule systems rely on per-sample evaluation, causing costs to scale with dataset size and introducing stochastic, hallucination-prone outputs. We propose reframing LLMs as \textit{code generators} rather than per-instance evaluators. A single LLM call generates executable, human-readable decision logic that runs deterministically over structured data, eliminating per-sample LLM queries while enabling reproducible and auditable predictions. We combine code generation with automated statistical validation using precision lift, binomial significance testing, and coverage filtering, and apply cluster-based gap analysis to iteratively refine decision logic without human annotation. We instantiate this framework in venture capital founder screening, a rare-event prediction task with strong interpretability requirements. On VCBench, a benchmark of 4,500 founders with a 9\% base success rate, our approach achieves 37.5\% precision and  an F$_{0.5}$ score of 25.0\%, outperforming GPT-4o (at 30.0\% precision and an F$_{0.5}$ score of 25.7\%) while maintaining full interpretability. Each prediction traces to executable rules over human-readable attributes, demonstrating verifiable and interpretable LLM-based decision-making in practice.
\end{abstract}


\section{Introduction}

The venture capital industry deploys over \$300 billion annually, yet investment decisions remain largely driven by intuition and pattern-matching from experience. A central challenge is founder evaluation: given a candidate's background, including their education, career history, and prior ventures, can we systematically identify patterns that predict startup success? The appeal of automation is clear, but the stakes demand transparency. When a VC passes on a founder or presents a deal to partners, they need to articulate \textit{why}.

This creates a tension between two desirable properties. \textbf{Interpretability} requires that predictions decompose into human-readable factors, such as "this founder has a prior exit and held a VP role at a scaled company," that investors can validate against their own judgment. \textbf{Scalability} requires screening thousands of candidates without proportional expert effort. Traditional approaches sacrifice one for the other: handcrafted rules are interpretable but labor-intensive, while black-box models scale but obscure their reasoning.

Recent work has turned to LLMs as a middle path, using them to generate human-readable screening rules~\cite{griffin2025rrf}. The idea is appealing: LLMs encode broad knowledge about what makes founders successful and can articulate it as natural language predicates. But these approaches evaluate each rule by querying the LLM for each founder. If a rule asks "Did this founder scale a previous company past 200 employees?", the LLM must answer that question 4,500 times. This leads to API costs that scale as O(rules $\times$ founders) and, more problematically, introduces stochastic variation where the same question about the same founder may receive different answers across runs.

We observe that this per-sample evaluation is unnecessary. Founder profiles are \textit{structured data}, consisting of education records, employment histories, and exit events, not free-form text requiring language understanding. A rule like "founder scaled a company past 200 employees" can be expressed as executable code that queries structured fields deterministically. This reframes the LLM's role: rather than serving as a rule \textit{evaluator} that is called repeatedly, it serves as a rule \textit{generator} that is called once.

This paper develops that insight into a complete framework. We prompt an LLM to generate Python predicates over founder attributes, validate rules automatically using precision lift and statistical significance, and apply cluster-based gap analysis to guide iterative refinement. On VCBench, our approach achieves 37.5\% precision and an F$_{0.5}$ 25.0\%, outperforming, GPT-4o, and other rivaling frontier LLMs (e.g, GPT-5, o3), while ensuring that every prediction traces to interpretable founder attributes.

We make the following contributions:
\begin{itemize}
    \item We propose treating LLMs as code generators rather than rule evaluators for founder screening. This architectural shift reduces API costs by over 99\% and eliminates per-sample hallucination risk while preserving interpretable outputs.
    \item We introduce automated statistical validation using precision lift, binomial significance testing, and coverage thresholds to filter low-quality rules without requiring human annotation.
    \item We develop a cluster-based gap analysis method that identifies founder subpopulations where existing rules fail to discriminate, providing structured feedback to guide iterative rule refinement.

\end{itemize}

\section{Related Work}

\subsection{LLM-Based Feature Engineering for Tabular Data}

Recent work has explored using LLMs to automate feature engineering for tabular prediction tasks. CAAFE \cite{hollmann2023caafe} introduces context-aware automated feature engineering, where an LLM iteratively generates Python code for new features based on dataset descriptions and cross-validation feedback. The system improved ROC AUC from 0.79 to 0.82 across 14 tabular datasets by leveraging the LLM's semantic understanding of feature relationships.

Several approaches extend this paradigm. OCTree \cite{nam2024decisiontree} uses decision tree reasoning to provide structured feedback to the LLM, conveying which features contribute to prediction errors and guiding the generation of corrective features. LLM-FE \cite{abhyankar2025llmfe} formulates feature engineering as a program search problem, combining evolutionary optimization with LLM-generated feature candidates. TabLLM \cite{tabllm2023fewshot} demonstrates that LLMs can engineer features for few-shot tabular learning by generating task-specific transformations from minimal examples.

These methods generate arbitrary feature transformations optimized for predictive performance, but the resulting features are often difficult for domain experts to interpret or validate.

\subsection{LLM-Based Feature Engineering for Interpretability}

A related line of work focuses specifically on generating interpretable features. FeatLLM \cite{balek2024llmtext} employs LLMs to produce binary features from text data for interpretable machine learning, emphasizing transparency over raw predictive performance. AutoQual \cite{autoqual2024} uses an LLM agent to discover interpretable features for review-quality assessment, iteratively refining feature definitions based on downstream task performance. GPTree \cite{xiong2024gptree} combines LLM-powered feature generation with decision tree classifiers to produce explainable predictions.

These approaches achieve interpretability but require the LLM to evaluate each rule on each sample, leading to costs that scale with dataset size and introducing stochastic variation across runs.

\subsection{Machine Learning for Venture Capital}

Predicting startup success has attracted significant research interest due to its economic importance and the challenge of learning from sparse, noisy signals. VCBench \cite{chen2025vcbenchbenchmarkingllmsventure} introduces the first standardized benchmark for founder success prediction, providing 4,500 anonymized founder profiles with standardized features covering education, employment history, and prior exits. The benchmark evaluates frontier LLMs directly on the prediction task, with GPT-5 achieving 59.0\% precision and an F$_{0.5}$ score of 16.2\%.

Random Rule Forest (RRF) \cite{griffin2025rrf} generates interpretable ensembles of LLM-generated questions, using the LLM to both propose and evaluate binary rules on founder profiles. Kumar et al. \cite{kumar2025rarevc} address the rare-event nature of startup success through LLM-powered feature engineering combined with multi-model learning.

\subsection{Positioning of Our Work}

Table \ref{tab:related_work} compares our approach with previous work across five dimensions that matter for practical deployment.

\begin{table}[h]
\centering
\caption{Comparison with related work on automated feature engineering across five key dimensions.}
\label{tab:related_work}
\begin{tabular}{lccccc}
\toprule
\textbf{Method} & \textbf{Cost} & \textbf{Speed} & \textbf{Interpret.} & \textbf{Stat. valid.} & \textbf{Determin.} \\
\midrule
CAAFE \cite{hollmann2023caafe} & Low & Fast & Low & CV only & Yes \\
FeatLLM \cite{balek2024llmtext} & High & Slow & High & None & No \\
RRF \cite{griffin2025rrf} & High & Slow & High & None & No \\
\textbf{Ours} & \textbf{Low} & \textbf{Fast} & \textbf{High} & \textbf{Built-in} & \textbf{Yes} \\
\bottomrule
\end{tabular}
\end{table}

We define these dimensions as follows:
\begin{itemize}
    \item \textbf{Cost}: Total API expense for rule generation and evaluation. Methods requiring per-sample LLM calls incur costs that grow with dataset size.
    \item \textbf{Speed}: Latency from input to prediction. Per-sample evaluation creates bottlenecks for large datasets.
    \item \textbf{Interpretability}: Whether predictions decompose into human-readable factors that domain experts can validate.
    \item \textbf{Statistical validation}: Whether the method includes built-in mechanisms to filter spurious patterns, beyond cross-validation on downstream tasks.
    \item \textbf{Deterministic outputs}: Whether the same input produces identical outputs across runs, enabling reproducibility and debugging.
\end{itemize}

Code generation methods like CAAFE execute efficiently but produce arbitrary feature transformations that lack interpretability. Interpretable methods like FeatLLM and RRF generate human-readable rules but require O(N $\times$ M) LLM calls to evaluate N rules on M samples, incurring high cost, slow speed, and stochastic outputs where the same rule may yield inconsistent evaluations.

Our approach combines the efficiency of code generation with the interpretability of rule-based methods. By prompting the LLM once to generate executable Python predicates, we achieve low cost, fast speed, and deterministic outputs. By generating binary rules over structured founder attributes, we maintain full interpretability. And by introducing precision lift, significance testing, and coverage thresholds, we provide built-in statistical validation that filters low-quality rules without human annotation.

%

\section{Methodology}

\begin{figure*}[t]
\centering
\includegraphics[width=\textwidth]{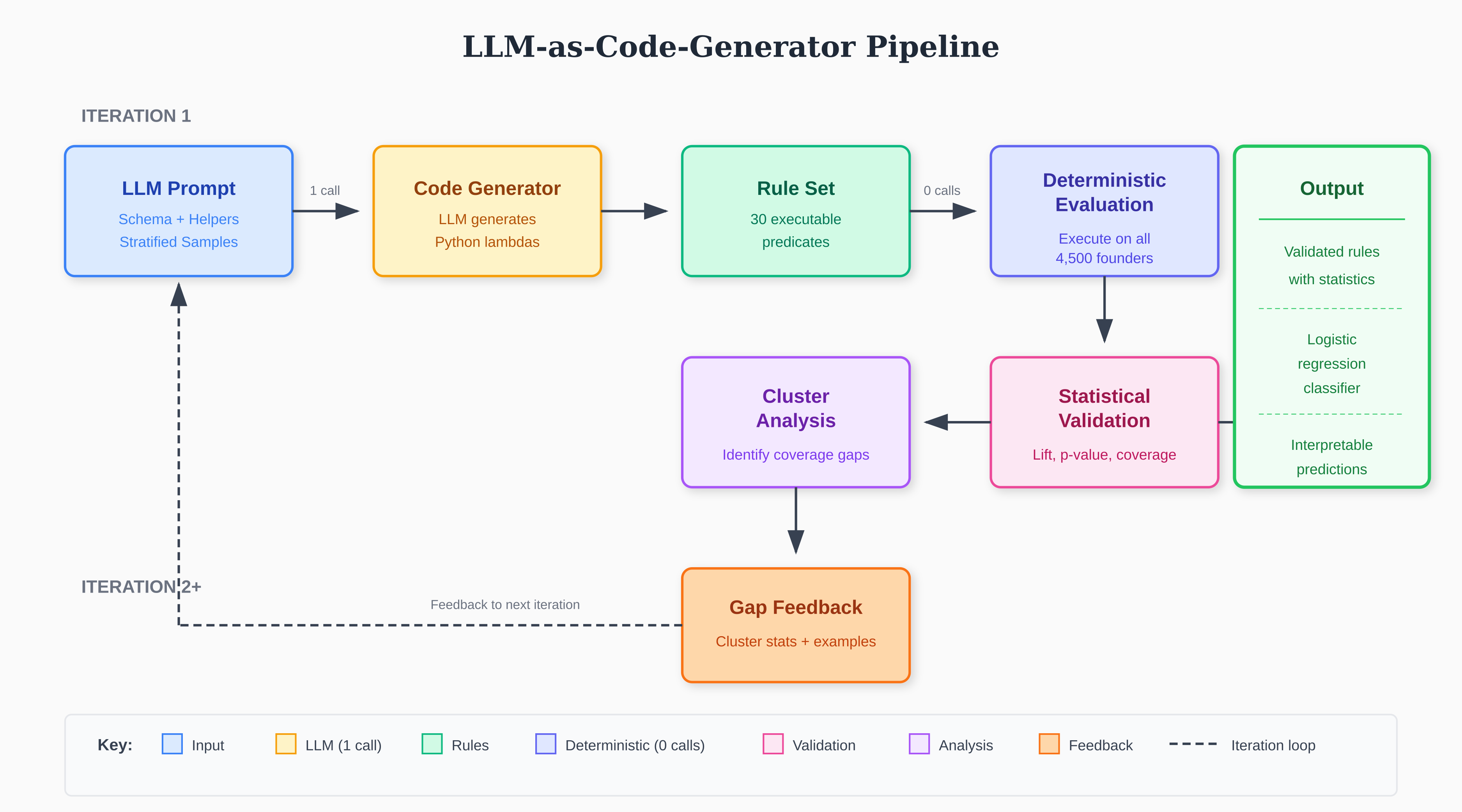}
\caption{Overview of our LLM-as-code-generator pipeline. The LLM is called once to generate executable Python rules, which then evaluate deterministically across all founders without further LLM involvement. Statistical validation filters low-quality rules, and cluster-based gap analysis guides iterative refinement.}
\label{fig:pipeline}
\end{figure*}

We present a framework for interpretable feature engineering that combines LLM-based code generation with statistical validation and cluster-guided iteration. Figure~\ref{fig:pipeline} illustrates the complete pipeline.

\subsection{Problem Setup}

Given a dataset of founder profiles with binary success labels, our goal is to generate a set of interpretable binary rules that predict founder success. Each rule should be human-readable, statistically predictive, and non-redundant with other rules. The final model is a logistic regression classifier trained on rule activations, enabling transparent predictions where each decision traces to specific founder attributes.

\subsection{Founder Data Representation}

Each founder profile is a structured record containing education history (degree type, field of study, university QS ranking), employment history (role titles, company sizes, industries, tenure durations), prior exits (IPOs and acquisitions from previous ventures), and the industry of the founder's current startup. Table~\ref{tab:founder_schema} summarizes the data structure.

\begin{table}[h]
\centering
\caption{Founder profile data structure.}
\label{tab:founder_schema}
\begin{tabular}{lll}
\toprule
\textbf{Field} & \textbf{Type} & \textbf{Example Values} \\
\midrule
industry & string & ``Software Development'' \\
educations & list & degree, field, qs\_ranking \\
jobs & list & role, company\_size, industry, duration \\
ipos & list & prior IPO events \\
acquisitions & list & prior acquisition events \\
\bottomrule
\end{tabular}
\end{table}

This structured representation enables the LLM to generate executable code that directly accesses founder attributes. We provide three helper functions that handle categorical parsing: \texttt{parse\_qs} converts QS ranking strings to numeric values, \texttt{parse\_duration} converts tenure strings to years, and \texttt{parse\_company\_size} converts company size ranges to employee counts. These helpers ensure that the LLM can generate syntactically valid predicates without implementing low-level parsing logic.

\subsection{Rule Generation via LLM Code Generation}

\begin{figure*}[t]
\centering
\includegraphics[width=0.95\textwidth]{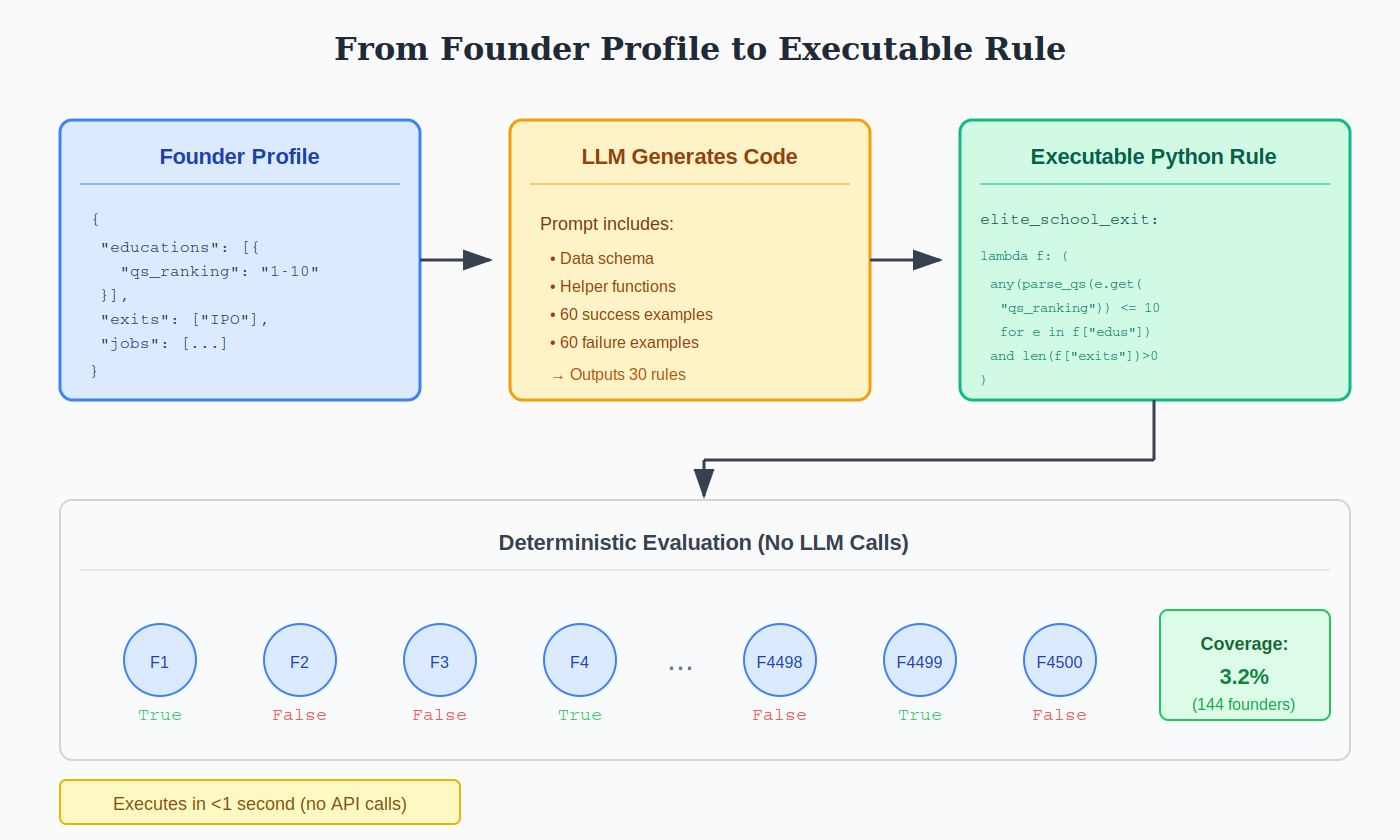}

\caption{Rule generation process. The LLM receives structured founder profiles and generates Python lambda expressions that encode binary screening rules. These rules execute deterministically across all 4,500 founders without additional LLM calls.}
\label{fig:rule_gen}
\end{figure*}

The core insight of our approach is that founder screening rules can be expressed as executable code rather than natural language predicates requiring LLM interpretation. Figure~\ref{fig:rule_gen} illustrates this process.

\subsubsection{Prompt Construction}

For each iteration, we construct a prompt with three components.

\textbf{Schema and helpers.} We provide the complete founder data structure and helper function signatures. We also include eight example rules demonstrating proper syntax, edge case handling, and diverse pattern types covering education, experience, exits, and industry matching.

\textbf{Stratified samples.} We include 120 founder profiles in each prompt, with 60 successful and 60 unsuccessful founders. This balanced sampling corrects for the dataset's 9\% base rate, preventing the LLM from generating rules that trivially predict failure. We select samples for diversity by bucketing founders by industry, education count, and job count, then sampling evenly across buckets. 

The sample size of 120 balances context window constraints against the need for sufficient diversity. In ablation experiments, we find that 50 samples yielded rules that were too specific: only 30\% passed the 1-50\% coverage threshold, with most rules matching fewer than 1\% of the full dataset (indicating overfitting to the limited prompt examples). Conversely, increasing to 200 samples showed negligible improvement in validation pass rate or mean precision lift while increasing prompt length significantly. The 120-sample configuration achieves high rule quality (63\% validation pass rate in Iteration 1) while remaining within practical prompt length limits, ensuring the LLM observes sufficient diversity to generate rules that generalize beyond the prompt sample.

\textbf{Prior iteration feedback.} For iterations beyond the first, we provide verbal reinforcement inspired by Reflexion~\cite{shinn2023reflexion}, which shows that LLMs can learn from trial-and-error through linguistic feedback rather than weight updates. We append previously generated rules with their precision lift, coverage, and statistical significance, along with cluster-based gap analysis (Section~\ref{sec:cluster}) that identifies founder subpopulations where existing rules fail to discriminate. Unlike Reflexion's binary success signals, our feedback is grounded in quantitative metrics, enabling the LLM to build on successful patterns, avoid redundancy, and target specific coverage gaps.

\subsubsection{Output Format}

The LLM generates rules as Python lambda expressions in a structured format:

\begin{verbatim}
1. rule_name ||| Description ||| lambda founder: <expr>
\end{verbatim}

For example, the rule \texttt{elite\_school\_and\_exit} checks whether a founder attended a top-50 QS-ranked university and has at least one prior exit:

\begin{verbatim}
lambda founder: (
    any(parse_qs(e.get("qs_ranking","")) <= 50 
        for e in founder.get("educations", [])) 
    and len(founder.get("ipos", []) or []) + 
        len(founder.get("acquisitions", []) or []) > 0)
\end{verbatim}

Each rule is a valid Python expression that takes a founder dictionary and returns True or False. Rules that fail to compile are discarded. We request 30 candidate rules per iteration, a quantity that balances exploration against redundancy: fewer rules reduce the likelihood of discovering high-lift patterns, while more rules yield diminishing returns as the LLM produces minor variations rather than novel patterns. Ablation experiments show that generating 50 rules yielded only 40\% validation pass rate, compared to 63\% for 30 rules indicating that the LLM generates lower-quality rules when forced to produce larger quantities. Beyond 30 rules, we observe diminishing returns as the model resorts to minor variations of existing patterns rather than discovering genuinely novel features.

\subsection{Deterministic Rule Evaluation}
Once generated, rules are compiled into callable Python functions and executed on all founders without additional LLM involvement. This evaluation strategy decouples cost from dataset size: regardless of whether we evaluate 1,000 or 100,000 founders, the API cost remains constant at one call per iteration. In contrast, per-sample evaluation approaches require one LLM call per (rule, founder) pair, causing cost to scale linearly with both the number of rules and dataset size, even with batched inference, total API usage grows proportionally with the number of founders.

Code execution provides two key advantages beyond cost independence. First, it ensures reproducibility: the same rule produces identical outputs across runs, eliminating the stochastic variation inherent in LLM-based evaluation where temperature and sampling can yield different answers to the same question. Second, it enables debugging: when a rule produces unexpected results, the Python code can be inspected, modified, and rerun, whereas LLM evaluation is opaque and cannot be systematically analyzed or corrected.

\subsection{Statistical Validation}

For each generated rule, we compute three metrics and retain only rules meeting all thresholds. Table~\ref{tab:validation} summarizes these criteria.

\begin{table}[h]
\centering
\caption{Statistical validation criteria for rule filtering.}
\label{tab:validation}
\begin{tabular}{lll}
\toprule
\textbf{Metric} & \textbf{Threshold} & \textbf{Rationale} \\
\midrule
Precision lift & $> 1.0$ & Above-baseline success rate \\
p-value & $< 0.05$ & Statistical significance \\
Coverage & 1\%--50\% & Sufficient power, meaningful discrimination \\
\bottomrule
\end{tabular}
\end{table}

\textbf{Precision lift.} We define precision lift as the ratio of success rate among matched founders to the baseline success rate:
\begin{equation}
\text{Precision Lift} = \frac{P(\text{success} \mid \text{rule matches})}{P(\text{success})}
\end{equation}
We require lift $> 1.0$, ensuring that the rules identify founders with above-baseline success rates. This threshold is deliberately permissive; stronger rules are naturally prioritized by the downstream logistic regression through learned coefficients.

\textbf{Statistical significance.} We test whether the observed success rate differs from baseline using a one-sided binomial test (normal approximation). We require $p < 0.05$, the conventional significance threshold. Given our iterative refinement and downstream cross-validation, we accept the 5\% false positive rate as rules receive additional validation.

\textbf{Coverage.} We define coverage as the fraction of founders matching the rule and require it to fall within [1\%, 50\%]. The minimum threshold (1\%, or 45 founders) ensures sufficient sample size for reliable significance testing. The maximum threshold (50\%) ensures rules provide meaningful discrimination rather than capturing generic attributes, and prevents multicollinearity in the downstream classifier.

\subsection{Cluster-Based Gap Analysis}
\label{sec:cluster}

Statistical validation identifies individually strong rules but does not guarantee collective coverage. A rule set might achieve high precision on founders it covers while missing entire subpopulations with different success patterns. To address this, we cluster founders based on their rule activation patterns using K-means, selecting the number of clusters via the elbow method (typically 4-5 clusters).

For each cluster, we compute success rate and classify it into one of three types:

\textbf{High-success clusters} (success rate $>$ 15\%) validate that existing rules are capturing genuine signal. These require no intervention.

\textbf{Low-success clusters} (success rate $<$ 7\%, size $>$ 400) are large populations where few founders succeed. However, the rare winners in these clusters may share characteristics that current rules miss.

\textbf{Mixed clusters} (success rate 7-12\%, size $>$ 400) indicate subpopulations where current rules fail to discriminate between success and failure. Founders in these clusters have similar rule activation patterns but different outcomes.

We also identify \textbf{uncovered successes}: successful founders who do not match any of the top rules, representing alternative success patterns not yet captured.

For each gap, we generate structured feedback including cluster statistics, example founders (both successful and unsuccessful), distinctive rule patterns for the cluster, and specific task instructions. For mixed clusters, we ask: ``What differentiates success from failure here?'' For low-success clusters with rare winners, we ask: ``What makes these rare successes different?'' This targeted feedback guides the next iteration toward rules that address identified weaknesses.

\subsection{Downstream Classification}

The validated rules form a binary feature matrix where each row is a founder and each column indicates whether a rule matches. We train a logistic regression classifier with balanced class weights using 5-fold stratified cross-validation.

We tune the classification threshold to optimize F$_{0.5}$, which weights precision higher than recall. In VC screening, high recall with low precision would overwhelm decision-makers with false positives, whereas high precision yields a manageable, high-confidence shortlist. The logistic regression coefficients provide interpretable importance weights, and predictions can be explained by listing the active rules for a given founder.

\subsection{Iterative Pipeline}

The complete pipeline proceeds as follows:

\textbf{Iteration 1:} Generate 30 rules without feedback, evaluate deterministically on all founders, compute precision lift and significance, filter by validation criteria, train logistic regression.

\textbf{Iteration 2+:} Cluster founders on current rule activations, identify gaps (mixed clusters, uncovered successes), generate feedback prompt with gap analysis and example founders, generate 30 new rules targeting gaps, evaluate, validate, and retrain.

Each iteration produces 30 candidate rules, of which 60-70\% typically pass statistical validation. Later iterations target gaps identified in earlier iterations, improving coverage and diversity of the rule set.

\section{Experiments}

We evaluate our framework on VCBench~\cite{chen2025vcbenchbenchmarkingllmsventure}, a standardized benchmark for founder success prediction.

\subsection{Setup}

\textbf{Dataset.} VCBench provides 4,500 anonymized founder profiles with binary success labels (IPO, acquisition, or fundraising $>$\$500M). The baseline success rate is 9.0\%, reflecting the rare-event nature of startup success.

\textbf{Implementation.} We use GPT-5.2 for rule generation, generating 30 rules per iteration across 2 iterations. The rules are filtered by precision lift $>$1.0, p-value $<$0.05, and coverage 1-50\%. The downstream classifier is logistic regression with balanced class weights, evaluated via 5-fold stratified cross-validation with threshold tuned for F$_{0.5}$.

\textbf{Baselines.} We compare against frontier LLMs evaluated directly on VCBench (GPT-5, GPT-4o, o3, Claude-3.5-Haiku, Gemini-2.5-Pro), human experts (Y Combinator at 14.0\% precision, Tier-1 VCs at 23.0\%), and the random baseline (9.0\%).

\subsection{Main Results}

Table~\ref{tab:main_results} presents detailed results.

\begin{table}[h]
\centering
\caption{Performance comparison on VCBench.}
\label{tab:main_results}
\begin{tabular}{lccc}
\toprule
\textbf{Method} & \textbf{Precision} & \textbf{Recall} & \textbf{F$_{0.5}$} \\
\midrule
Random Baseline & 9.0\% & 9.0\% & 9.0\% \\
Y Combinator & 14.0\% & 6.9\% & 8.6\% \\
Tier-1 VCs & 23.0\% & 5.2\% & 10.7\% \\
\midrule
Claude-3.5-Haiku & 15.8\% & 46.4\% & 18.2\% \\
GPT-4o & 30.0\% & 16.3\% & 25.7\% \\
o3 & 43.2\% & 7.4\% & 21.5\% \\
\midrule
\textbf{Ours (Iter 1)} & 30.6\% & 21.0\% & 27.7\% \\
\textbf{Ours (Iter 2)} & \textbf{37.5\%} & 16.3\% & 25.0\% \\
\bottomrule
\end{tabular}
\end{table}

Our Iteration 1 model matches GPT-4o precision (30.6\% vs 30.0\%) with higher F$_{0.5}$ (27.7\% vs 25.7\%). After cluster-based feedback, Iteration 2 achieves 37.5\% precision, a 22\% relative improvement. Critically, unlike black-box LLM predictions, our results are fully interpretable.

\subsection{Rule Analysis}

\begin{figure}[h]
\centering
\includegraphics[width=0.9\columnwidth]{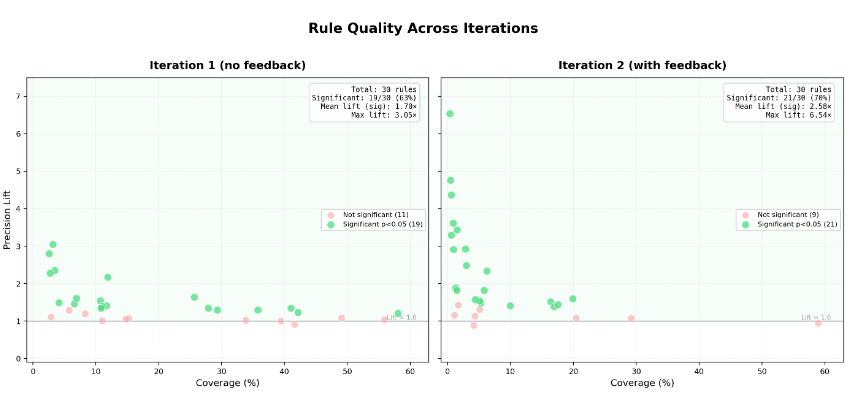}

\caption{Rule quality distribution. Each point is a generated rule. Significant rules (green) cluster above the lift=1.0 threshold, with a tradeoff between lift and coverage.}
\label{fig:rule_quality}
\end{figure}

Figure~\ref{fig:rule_quality} shows the distribution of generated rules by precision lift and coverage. Green points indicate statistically significant rules (p$<$0.05).

\begin{table}[h]
\centering
\caption{Top rules by precision lift per iteration.}
\label{tab:top_rules}
\begin{tabular}{llcc}
\toprule
\textbf{Iter} & \textbf{Rule} & \textbf{Lift} & \textbf{Cov.} \\
\midrule
1 & prior\_exit & 3.05$\times$ & 3.16\% \\
1 & founder\_scaled\_company\_200plus & 2.80$\times$ & 2.56\% \\
1 & worked\_in\_vc\_pe & 2.35$\times$ & 3.47\% \\
1 & executive\_chair\_or\_operating\_partner & 2.28$\times$ & 2.71\% \\
1 & elite\_qs\_top10 & 2.18$\times$ & 11.91\% \\
\midrule
2 & cybersecurity\_domain\_depth & 6.54$\times$ & 0.38\% \\
2 & ipo\_only\_exit & 4.76$\times$ & 0.47\% \\
2 & repeat\_exit\_founder & 4.37$\times$ & 0.62\% \\
2 & board\_plus\_exit & 3.61$\times$ & 0.89\% \\
2 & founder\_scaled\_500plus & 3.43$\times$ & 1.51\% \\
\bottomrule
\end{tabular}
\end{table}

Iteration 1 discovers signals around exits, elite credentials, and domain expertise. Iteration 2, guided by gap analysis, discovers complementary patterns like repeat entrepreneurship and finance backgrounds.

\subsection{Iteration Improvement}

Cluster analysis reveals that 68\% of founders fell into mixed clusters where Iteration 1 rules failed to discriminate, and 57\% of successful founders were not covered by the top rules. This feedback guides Iteration 2 toward specialized rules targeting these gaps as seen in in Table~\ref{tab:top_rules}.

\subsection{Cost Analysis}

Our approach requires one LLM call for rule generation; all evaluation is deterministic code execution. LLM-based evaluation would require multiple calls in proportion to the number of rules and founders, representing a 99\% reduction with corresponding cost savings and elimination of per-sample hallucination risk.

\section{Discussion}

\subsection{Interpretability, Performance, and the Advantages of Code-Based Evaluation}

Our results show that interpretable rule-based models can compete with black-box LLMs on founder success prediction. At 37.5\% precision and an F$_{0.5}$ score of 25.0\%, our approach outperforms GPT-4o (30.0\% precision, F$_{0.5}$ = 25.7\%) while providing full transparency. Although o3 and GPT-5 achieve higher precision (43.2\% and 59.1\% respectively), their F$_{0.5}$ scores (21.5\% and 16.1\%) fall well below ours, reflecting operating points that sacrifice recall for narrow, high-confidence predictions. Our method achieves a balanced Precision and F$_{0.5}$ among the models compared.

high-precision models tend to concentrate on a narrow founder archetype, starving deal flow and typically missing the non-obvious outliers that drive returns. Worse, black-box LLMs provide no visibility into what that archetype actually is, making it difficult to evaluate whether the model’s selectivity reflects genuine insight or bias. F$_{0.5}$ captures the precision-recall tradeoff relevant to this setting, and our method achieves the strongest F$_{0.5}$ score among interpretable approaches. 

Our method provides three advantages beyond cost reduction:

\textbf{Determinism.} The same rule produces identical results across runs, enabling reproducible analysis. LLM evaluation introduces stochastic variation where identical queries may yield different answers.

\textbf{Scalability.} Code execution scales to any dataset size without API costs. We evaluate 30 rules on 4,500 founders in seconds; LLM evaluation would require hours and significant expense.

\textbf{Debuggability.} Rule-based evaluation systems are transparent and interpretable, allowing decisions to be traced directly to the logic and input features that produced them. This enables effective debugging, reproducibility, and systematic refinement. In contrast, LLM-based evaluations are largely opaque, offering limited insight into the reasoning behind individual outcomes.

This opacity also complicates bias detection and auditing. Code-based rules provide explicit feature attribution, making it possible to assess whether particular inputs influenced a decision. Black-box LLM outputs lack such traceability, making fairness evaluation, accountability, and governance more difficult in practice.

\subsection{Human-in-the-Loop Decision Making}

Interpretable rule-based systems naturally support human-in-the-loop workflows, allowing analysts to inspect, contextualize, and override model outputs in high-stakes settings such as VC screening. Because decisions are grounded in explicit rules and features, human feedback can be incorporated deterministically, enabling accountability, bias auditing, and systematic refinement in ways that are difficult with black-box LLM evaluations.

\subsection{Limitations}

\textbf{Rule expressiveness.} Our rules are limited to patterns expressible as Python predicates over structured fields. Complex patterns requiring natural language understanding of job descriptions may need hybrid approaches.

\textbf{Coverage-precision tradeoff.} Iteration 2 improves precision but decreases recall as the LLM generated more specialized rules. Balancing this tradeoff may require explicit coverage targets or multi-objective optimization.

\textbf{Dataset specificity.} Evaluation is limited to VCBench. Generalization to other datasets or success definitions requires further validation.

\textbf{Cluster-guided rule refinement.} We use clustering over generated binary rules to organize them by similarity, quantified clusters by precision lift and prediction frequency, and qualitatively inspected rule variation across clusters to guide subsequent LLM prompts. While this process proved effective for iterative improvement, it remains a heuristic, analyst-in-the-loop approach rather than a fully automated or theoretically grounded optimization procedure.

\subsection{Future Work}

\textbf{Reinforcement learning from verifiable rewards.} Statistical validation metrics provide automatic reward signals. Fine-tuning LLMs on successful rules via reinforcement learning could improve generation quality without human annotation.

\textbf{Cross-domain application.} The code generation and validation approach is domain-agnostic. Applying this framework to other tabular prediction tasks (loan default, medical diagnosis, customer churn) would test generalization.

\textbf{Rule combination optimization.} We use logistic regression to combine rules. More sophisticated methods such as rule set learning or weighted voting could improve performance while maintaining interpretability.

\section{Conclusion}

We present a framework for interpretable feature engineering that uses LLMs to generate executable Python rules rather than evaluate samples directly. This paradigm shift reduces API costs by 99\%, eliminates per-sample hallucination risk, and produces deterministic, debuggable predictions. Statistical validation provides automatic quality filtering without human annotation, and cluster-based gap analysis guides iterative refinement.

On VCBench, our approach achieves 37.5\% precision and an F$_{0.5}$ score of 25\%, outperforming GPT-4o and other rival frontier LLMs (e.g., GPT-5, o3), while ensuring that every prediction is traceable to interpretable founder attributes.

Our work demonstrates that LLMs can serve as effective hypothesis generators when paired with statistical validation and structured feedback. The generated rules are not black-box predictions but interpretable features that domain experts can inspect, validate, and refine, combining the scalability of automation with the transparency of manual feature engineering.

\section*{Acknowledgements}
This work was done outside of the current role that Yagiz Ihlamur holds at Amazon.

\bibliographystyle{unsrt}
\bibliography{references}

\appendix

\section{Prompt Template}
\label{app:prompt}

The following is an abbreviated version of the system prompt used for rule generation. The full prompt includes eight example rules and detailed formatting instructions.

\begin{small}
\begin{verbatim}
You are an expert feature engineer. Generate binary
classification rules to predict startup founder success.

## TASK
Generate exactly 30 binary rules. Each rule MUST be a
valid Python lambda expression that takes a `founder`
dict and returns True or False.

## OUTPUT FORMAT
1. rule_name ||| Description ||| lambda founder: <expr>

## FOUNDER DATA STRUCTURE
founder = {
    "industry": str,
    "educations": [
        {"degree": str, "field": str,
         "qs_ranking": str}
    ],
    "jobs": [
        {"role": str, "company_size": str,
         "industry": str, "duration": str}
    ],
    "ipos": list,
    "acquisitions": list
}

## HELPER FUNCTIONS
parse_qs(qs_str) -> float
parse_duration(dur_str) -> float
parse_company_size(size_str) -> int

## RULES FOR WRITING EXPRESSIONS
1. Handle empty lists with any(), len()
2. Use .get() for dict access with defaults
3. Use .lower() for string comparisons
4. Return boolean (True/False)
\end{verbatim}
\end{small}

For iterations beyond the first, the prompt is augmented with a feedback section containing all previously generated rules annotated with their precision lift, coverage, p-value, and significance status, followed by cluster-based gap analysis output (see Section~\ref{sec:cluster}).

\section{Founder Data Example}
\label{app:founder}

An example founder profile from VCBench (anonymized):

\begin{small}
\begin{verbatim}
{
  "industry": "Software Development",
  "educations": [
    {"degree": "MS",
     "field": "Computer Science",
     "qs_ranking": "15"}
  ],
  "jobs": [
    {"role": "CTO",
     "company_size": "51-200",
     "industry": "Software",
     "duration": "4-5"},
    {"role": "Senior Engineer",
     "company_size": "10001+",
     "industry": "Technology",
     "duration": "2-3"}
  ],
  "ipos": [],
  "acquisitions": [{"company": "StartupCo"}]
}
\end{verbatim}
\end{small}

\section{Example Generated Rules}
\label{app:rules}

Table~\ref{tab:example_rules} shows representative rules from each iteration with their executable code and validation statistics. All expressions are reproduced verbatim from the pipeline output.

\begin{table*}[h]
\centering
\caption{Representative generated rules with validation statistics. Iter~1 rules were generated without feedback; Iter~2 rules were guided by cluster-based gap analysis targeting underperforming subpopulations.}
\label{tab:example_rules}
\small
\begin{tabular}{@{}llcrr@{}}
\toprule
\textbf{Rule Name} & \textbf{Description} & \textbf{Lift} & \textbf{Coverage} & \textbf{p-value} \\
\midrule
\multicolumn{5}{@{}l}{\textit{Iteration 1: Initial generation (no feedback)}} \\[3pt]
\texttt{prior\_exit} & Founder has $\geq$1 prior IPO or acquisition & 3.05$\times$ & 3.16\% & $<$.001 \\[2pt]
\texttt{founder\_scaled\_200plus} & Founded/co-founded company that grew to 200+ employees & 2.80$\times$ & 2.56\% & $<$.001 \\[2pt]
\texttt{elite\_qs\_top10} & Attended top-10 QS-ranked university & 2.18$\times$ & 11.91\% & $<$.001 \\[6pt]
\multicolumn{5}{@{}l}{\textit{Iteration 2: Gap-targeted generation (with feedback)}} \\[3pt]
\texttt{cybersecurity\_domain\_depth} & Security industry + technical security role experience & 6.54$\times$ & 0.38\% & $<$.001 \\[2pt]
\texttt{repeat\_exit\_founder} & Founder has $\geq$2 prior exits (IPO or acquisition) & 4.37$\times$ & 0.62\% & $<$.001 \\[2pt]
\texttt{board\_plus\_exit} & Board/advisory experience combined with prior exit & 3.61$\times$ & 0.89\% & $<$.001 \\
\bottomrule
\end{tabular}

\vspace{6pt}
\end{table*}

\section{Complete Iteration Statistics}
\label{app:iter_stats}

Table~\ref{tab:iter_comparison} compares rule generation quality across iterations.

\begin{table}[h]
\centering
\caption{Rule generation statistics per iteration. Feedback-guided generation (Iter~2) produces more significant rules with higher mean precision lift, though with lower average coverage as the LLM targets more specialized subpopulations.}
\label{tab:iter_comparison}
\begin{tabular}{lcc}
\toprule
\textbf{Metric} & \textbf{Iter 1} & \textbf{Iter 2} \\
\midrule
Rules generated & 30 & 30 \\
Significant (p $<$ 0.05) & 19 (63\%) & 21 (70\%) \\
Mean lift (sig.\ rules) & 1.70$\times$ & 2.58$\times$ \\
Median lift (sig.\ rules) & 1.46$\times$ & 1.88$\times$ \\
Rules with lift $>$ 2$\times$ & 5 & 10 \\
Rules with lift $>$ 3$\times$ & 1 & 6 \\
Mean coverage (sig.\ rules) & 18.2\% & 5.8\% \\
Median coverage (sig.\ rules) & 11.0\% & 3.0\% \\
\bottomrule
\end{tabular}
\end{table}

\end{document}